\pdfoutput=1

\documentclass[11pt]{article}

\usepackage{microtype}
\usepackage{graphicx}
\usepackage{subfigure}
\usepackage{booktabs} 
\usepackage{graphicx}
\usepackage{hyperref}

\usepackage[preprint]{acl}

\usepackage{amsmath}
\usepackage{amssymb}
\usepackage{mathtools}
\usepackage{amsthm}

\usepackage{times}
\usepackage{latexsym}

\usepackage[T1]{fontenc}

\usepackage[utf8]{inputenc}

\usepackage{microtype}

\usepackage{inconsolata}

\usepackage{graphicx} 
\usepackage{luatex85}
\usepackage{threeparttable}
\usepackage{amsmath}
\usepackage{amsfonts}
\usepackage{bm}
\usepackage{subcaption}

\usepackage{inconsolata}
\usepackage{algorithm}
\usepackage{algorithmic}
\usepackage{xcolor}
\usepackage{tcolorbox}
\tcbuselibrary{raster} 
\definecolor{darkblue}{rgb}{0.04, 0.1, 0.35}
\definecolor{lightblue}{rgb}{0.2, 0.6, 0.9}
\definecolor{lightred}{rgb}{0.9, 0.4, 0.4}
\definecolor{lightyellow}{rgb}{0.8, 0.8, 0.35}
\definecolor{lightpurple}{rgb}{0.6, 0.5, 0.7}
\definecolor{lightorange}{rgb}{1.0, 0.5, 0.0}
\definecolor{lightgrey}{rgb}{0.5, 0.5, 0.5}
\usepackage{booktabs} 
\makeatletter
\newcommand*\bigcdot{\mathpalette\bigcdot@{.5}}
\newcommand*\bigcdot@[2]{\mathbin{\vcenter{\hbox{\scalebox{#2}{$\m@th#1\bullet$}}}}}
\makeatother
\usepackage[capitalize,noabbrev]{cleveref}

\theoremstyle{plain}
\newtheorem{theorem}{Theorem}[section]

\theoremstyle{definition}

\theoremstyle{remark}
\newtheorem{remark}[theorem]{Remark}

\newcommand{\cT}{\mathcal{T}}

\newcommand{\cX}{\mathcal{X}}
\newcommand{\cY}{\mathcal{Y}}
\newcommand{\cA}{\mathcal{A}}
\newcommand{\cS}{\mathcal{S}}

\newcommand{\cL}{\mathcal{L}}

\usepackage{times}
\usepackage{latexsym}

\usepackage{microtype}

\usepackage{inconsolata}

\usepackage{graphicx}
\usepackage{tcolorbox}
\title{ \textsc{FLAG-Trader}: Fusion LLM-Agent with Gradient-based \\Reinforcement Learning for
Financial Trading}

\author{%
Guojun Xiong\textsuperscript{1},
Zhiyang Deng\textsuperscript{2},
Keyi Wang\textsuperscript{3},
Yupeng Cao\textsuperscript{2},
Haohang Li\textsuperscript{2},
Yangyang Yu\textsuperscript{2},
\\
\textbf{Xueqing Peng\textsuperscript{7}},
\textbf{Mingquan Lin\textsuperscript{4}},
\textbf{Kaleb E Smith\textsuperscript{5}},
\textbf{Xiao-Yang Liu Yanglet\textsuperscript{3,6}},
\\
\textbf{Jimin Huang\textsuperscript{7}},
\textbf{Sophia Ananiadou\textsuperscript{8}},
\textbf{Qianqian Xie\textsuperscript{7,*}}
\\
\\
\textsuperscript{1}Harvard University,
\textsuperscript{2}Stevens Institute of Technology,
\textsuperscript{3}Columbia University,
\\
\textsuperscript{4}University of Minnesota,
\textsuperscript{5}NVIDIA,
\textsuperscript{6}Rensselaer Polytechnic Institute,
\\
\textsuperscript{7}TheFinAI,
\textsuperscript{8}University of Manchester
\\
\small{
\textbf{\textsuperscript{*}Correspondence:} \href{xqq.sincere@gmail.com}{xqq.sincere@gmail.com} 
}
}

\begin{document}
\maketitle
\begin{abstract}
Large language models (LLMs) fine-tuned on multimodal financial data have demonstrated impressive reasoning capabilities in various financial tasks. However, they often struggle with multi-step, goal-oriented scenarios in interactive financial markets, such as trading, where complex agentic approaches are required to improve decision-making. To address this, we propose \textsc{FLAG-Trader}, a unified architecture integrating linguistic processing (via LLMs) with gradient-driven reinforcement learning (RL) policy optimization, in which a partially fine-tuned LLM acts as the policy network, leveraging pre-trained knowledge while adapting to the financial domain through parameter-efficient fine-tuning.  Through policy gradient optimization driven by trading rewards, our framework not only enhances LLM performance in trading but also improves results on other financial-domain tasks. We present extensive empirical evidence to validate these enhancements.
\end{abstract}

\section{Introduction}

\begin{figure*}
     \centering
     \includegraphics[width=0.9\linewidth]{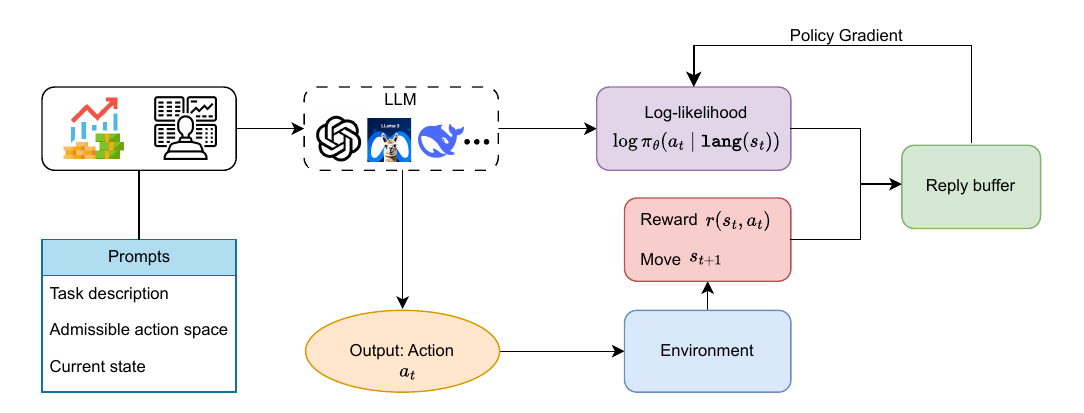}
     \caption{A high-level overview of our LLM-based reinforcement learning setup for financial trading. The environment provides the current state 
$s_t$. A prompt containing task details, the action space, and the current state is fed into the LLM, which outputs a trading action 
$a_t$. The action is executed in the environment, yielding a reward 
$r(s_t, a_t)$ and next state 
$s_{t+1}$. The log-likelihood 
$\log_{\pi_\theta}(a_t|\texttt{lang}(s_t))$ is then leveraged by a policy gradient method (e.g., PPO), with experience tuples stored in a replay buffer for iterative updates.}
     \label{fig:FinRL_LLM}
 \end{figure*}
Algorithmic financial trading represents a critically complex decision-making domain that perpetually grapples with the intertwined challenges of synthesizing heterogeneous market signals and dynamically refining strategies \cite{hambly2023recent,yu2024fincon,li2023tradinggpt}. Traditional reinforcement learning (RL) approaches, despite their theoretical grounding in Markov Decision Processes (MDPs), confront three fundamental limitations when deployed in financial markets. Firstly, their inability to coherently model multimodal market states—spanning sequential price movements, quantitative technical indicators, and unstructured textual sentiments—compromises data integration \cite{Zhang2019DeepRL,nassirtoussi2014text}. Secondly, non-stationary data distributions inherent to financial systems systematically erode strategy generalizability across market regimes \cite{Zhang2019DeepRL}. Thirdly, the heavy reliance on manually crafted technical indicators (e.g., MACD, RSI) and complex feature engineering \cite{liang2018adversarial} introduces subjective biases, leads to information loss, and reduces the robustness of real-time decision-making, especially in volatile market conditions.

The emergence of Large Language Models (LLMs) offer significant potential for financial decision-making by addressing key limitations of RL-based trading strategies. Leveraging their transformer architecture, they serve as multimodal feature extractors, integrating time-series and textual data, capturing long-range dependencies, and generalizing across market regimes, while also extracting nuanced sentiment signals without relying on manually crafted features \cite{chen2021decision,yang2023fingpt,jin2023time,wood2021trading,yu2024finmem,deng2023llms}. Nonetheless, adapting LLMs for trading presents key challenges. First, their deployment often relies on agentic frameworks \cite{li2024cryptotrade,li2023tradinggpt,yu2025fincon}, which incur high implementation and operational costs due to their complex architecture. Second, LLMs are primarily trained for static text generation, making them ill-suited for sequential decision-making in trading.  This prompts us to the following question:


\vspace{-0.1in}
\begin{tcolorbox}[colback=white!5!white,colframe=white!75!white]
\textit{
Can we design a framework that seamlessly integrates LLMs' reasoning with RL's reward-driven optimization to tackle the challenges of financial sequential decision-making? }
\end{tcolorbox}
\vspace{-0.15in}

To resolve these interconnected challenges, we \textsc{FLAG-Trader}, a unified architecture integrating linguistic processing (via LLMs) with gradient-driven RL policy optimization, as shown in Figure \ref{fig:FinRL_LLM}. This framework advances two synergistic innovations: a parameter-efficient fine-tuning module that jointly encodes temporal market data and textual streams into unified state representations and a hybrid RL component that explicitly incorporates external environment reward gradients into policy updates, ensuring alignment with trading performance metrics.  Our contributions are summarized as follows.

First, we propose the \textsc{FLAG-Trader} framework, where a partially fine-tuned LLM acts as the policy network, leveraging pre-trained knowledge while adapting to the financial domain through parameter-efficient fine-tuning. The model processes market data using a textual state representation, enabling it to interpret and respond to market conditions effectively. Rather than fine-tuning the entire LLM, only a subset of its parameters is updated, balancing domain adaptation and knowledge retention. This design allows \textsc{FLAG-Trader} to make informed trading decisions while remaining computationally efficient and preserving the LLM’s general reasoning capabilities.

Second, we conduct extensive experiments to evaluate \textsc{FLAG-Trader} across multiple financial trading tasks. Our results demonstrate that \textsc{FLAG-Trader} consistently outperforms both the buy-and-hold strategy and LLM-agentic baselines, particularly in terms of cumulative return and Sharpe ratio, which we prioritize for financial performance assessment. Notably, our approach enables a small-scale (135M parameter) open-source LLM to surpass much larger proprietary models, highlighting the effectiveness of RL fine-tuning in optimizing LLM-driven trading strategies. These findings underscore the potential of integrating LLMs with RL to enhance financial decision-making while maintaining computational efficiency.

\section{Related Work}
\label{sec:related-work}

\textbf{RL in Finance.}
RL has shown promise for financial decision-making, spanning Q-learning approaches for Sharpe ratio maximization \cite{gao2000algorithm}, dynamic asset allocation \cite{jangmin2006adaptive}, deep Q-learning \cite{jeong2019improving}, tabular SARSA \cite{de2020tabular}, policy-based portfolio optimization \cite{shi2019multi}, and actor-critic methods \cite{ye2020reinforcement} enhanced by adversarial training \cite{liang2018adversarial} and transformer-based architectures \cite{huang2024novel}. 
Recent research efforts in RL for financial applications have been greatly aided by open-source frameworks like FinRL \cite{liu2022finrl}, which standardize implementations and provide reproducible benchmarks. Comprehensive surveys \cite{hambly2023recent,sun2023reinforcement} further showcase advances in both methodological rigor and real-world deployment.
Despite these advances, RL-based trading still requires large training data, struggles with non-stationary markets, and faces challenges incorporating multimodal information in real time.

\textbf{LLMs in Finance.}
A growing trend is the integration of LLMs into financial decision-making. Hybrid systems like FinCon \cite{yu2025fincon} and TradingGPT \cite{li2023tradinggpt} leverage language understanding to enhance trading agents, while domain-specific models such as \textsc{FinBERT} \cite{araci2019finbert,yang2020finbert}, \textsc{FLANG} \cite{shah2022flue} and \textsc{Open-FinLLMs} \cite{xie2024open} have excelled at financial text tasks through specialized pre-training. Recent efforts include machine reading comprehension \cite{zhang2023finbert}, open-source financial LLMs \cite{liu2023fingpt}, BloombergGPT with domain-adapted tokenization \cite{wu2023bloomberggpt}, and InvestLM \cite{yang2023investlm} featuring numerical reasoning—achieving strong results in sentiment analysis \cite{huang2023finbert}, earnings call interpretation \cite{xie2023pixiu}, and regulatory document processing. Additionally, \textsc{FinBen} \cite{xie2024finben}, benchmark study for LLMs in finance, have emerged to comprehensively evaluate model performance across various financial tasks. However, LLM-based methods often lack sequential decision-making mechanisms, are computationally expensive (especially with RL), and struggle with non-stationary market conditions.

\textbf{LLM Agents for Sequential Decision Making.}
The integration of LLMs with agentic frameworks has opened new avenues for financial decision-making. For instance, \textsc{FinMem} \cite{yu2024finmem} introduced memory-augmented LLM agents for portfolio management, \textsc{FinAgent} \cite{zhang2024finagent} leveraged hierarchical structures in high-frequency trading, and multi-agent systems like \textsc{FinRobot} \cite{yang2024finrobot} and \textsc{FinCon} \cite{yu2024fincon} emphasize contextual adaptation and collaboration. Meanwhile, fine-tuning LLMs and vision-language models (VLMs) with reinforcement learning has proven effective in complex tasks: LLaRP \cite{szot2023large} positions LLMs as generalizable policies for embodied tasks, and RL-tuned VLMs \cite{zhai2024fine} enhance multi-step decision-making.
However, LLMs remain computationally expensive for real-time deployment, and risk-sensitive trading demands robustness to non-stationary markets, calling for careful model complexity and balanced exploration-exploitation.

 \begin{figure*}
     \centering
     \includegraphics[width=0.9\linewidth]{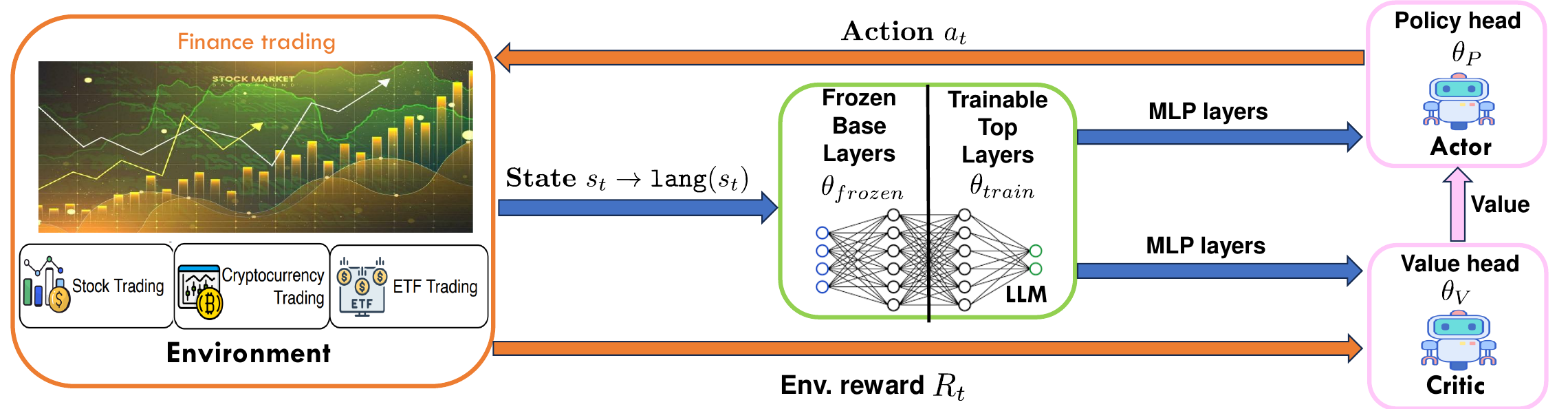}
     \caption{The \textsc{FLAG-Trader} pipeline for financial trading, utilizing an LLM-based actor-critic architecture.  The LLM consists of \textbf{frozen base layers} $\theta_{\texttt{frozen}}$ that retain pre-trained knowledge and \textbf{trainable top layers} $\theta_{\texttt{train}}$ for financial decision-making. Both the \textsc{Policy\_Net}  and \textsc{Value\_Net}  share these trainable layers while maintaining separate \textit{policy head} $\theta_P$ and \textit{value head} $\theta_V$, which are updated by policy gradient method.}
     \label{fig:AC-network}
 \end{figure*}
\section{Problem Statement}\label{Sec:Problem Formulation}

We define the financial decision-making process as a finite horizon partially observable Markov decision process (MDP) with time index $\{0,\cdots,T\}$, represented by the tuple:
$\mathcal{M} = (\mathcal{S}, \mathcal{A}, \mathcal{T}, R, \gamma),$
where each component is described in detail below.

\textbf{State.} The state space $\cS=\cX\times\cY$ consists of two components: market observations and trading account balance, i.e.,
$s_t = (m_t, b_t) \in \mathcal{S}.$ Specifically,
     $ m_t = (P_t, N_t) \in \cX$ represents the \textit{market observation process}, includingstock price $P_t$ at time $t$, and financial news sentiment or macroeconomic indicators $N_t$;
   $ b_t = (C_t, H_t) \in \mathcal{Y}$ represents the \textit{trading account balance}, including available cash $C_t$ at time $t$, and number of stock shares $H_t$.

\textbf{Action.} The agent chooses from a discrete set of trading actions
$\mathcal{A} = \{\texttt{Sell}: -1, \texttt{Hold}: 0, \texttt{Buy}: 1\},$
where $a_t=-1$ denotes selling all holdings (liquidate the portfolio),
$a_t=0$ denotes holding (no trading action), and
$a_t=1$ represents buying with all available cash (convert all cash into stocks).

 \textbf{State Transition.} The state transition dynamics are governed by a stochastic process
$s_{t+1} \sim \cT(\cdot | s_t, a_t).$
The trading account evolves according to the following equations:
    \begin{itemize}
        \item If \texttt{Sell}:
        $        C_{t+1} = C_t + H_t P_{t+1}, \quad H_{t+1} = 0.
        $
        \item If \texttt{Hold}:
        $ C_{t+1} = C_t, \quad H_{t+1} = H_t.$
        \item If \texttt{Buy}:
       $        C_{t+1} = 0, \quad H_{t+1} = H_t + \frac{C_t}{P_{t+1}}.$ 
    \end{itemize}

\textbf{Reward.} 
The agent receives a reward  based on the daily trading profit \& loss (PnLs):
\begin{align*}
    R(s_t, a_t) = SR_t-SR_{t-1},
\end{align*}
where $SR_t$ denotes the Sharpe ratio at day $t$, computed by using the historical PnL from time 0 to time $t$. Moreover, PnL at time $t$ is calculated as
\begin{align*}
pnl_t:=(C_t-C_{t-1})+(H_tP_t-H_{t-1}P_{t-1}).   
\end{align*}
Then, the Sharpe ratio $SR_t$ at time $t$ can be calculated as:
\begin{align}
    \label{eq:sharpe}
    SR_t:=\frac{\mathbb{E}[pnl_1,\cdots,pnl_t]-r_f}{\sigma[pnl_1,\cdots,pnl_t]},
\end{align}
where $\mathbb{E}[pnl_1,\cdots,pnl_t]$ is the sample average of daily PnL up to time $t$, $r_f$ is the risk-free rate, and $\sigma[pnl_1,\cdots,pnl_t]$ is the sample standard deviation of daily PnL up to time $t$.

The goal is to find an admissible policy $\pi$ to maximize the expected value of cumulative discounted reward, i.e., 
\begin{align}\label{eq:global_obj}
   \max_{\pi} V^{\pi}(s) = \mathop{\mathbb{E}}\limits_{\substack{s_0=s,a_t\sim \pi(\cdot|s_t)\\ s_{t+1}\sim \cT(\cdot|s_t,a_t)}} \left[ \sum_{t=0}^T \gamma^t R_t \right],
\end{align}
where $R_t$ is a shortened version $R(s_t, a_t)$ and  $ \gamma \in (0,1]$ is the discount factor controlling the importance of future rewards.

Our goal is to train an LLM agent parameterized by $\theta$ to find the optimized policy $\pi_\theta$ for \eqref{eq:global_obj}, i.e.,
\begin{align}
a_t\sim\pi_\theta(\cdot|s_t)=\text{LLM}(\texttt{lang}(s_t);\theta),
\end{align}
where $\texttt{lang}(s_t)$ are the prompts generated by converting state $s_t$ into structured text. The proposed pipeline is illustrated in Figure \ref{fig:FinRL_LLM}.

\section{\textsc{FLAG-Trader}}
To tackle the challenge of directly fine-tuning an LLM for both alignment and decision-making, we introduce \textsc{FLAG-Trader}, a fused LLM-agent and RL framework for financial stock trading. In \textsc{FLAG-Trader}, a partially fine-tuned LLM serves as the policy network, leveraging its pre-trained knowledge while adapting to the financial domain through parameter-efficient fine-tuning, as shown in Figure \ref{fig:AC-network}. The model processes financial information using a textual state representation, allowing it to interpret and respond to market conditions effectively. Instead of fine-tuning the entire network, only a subset of the LLM’s parameters is trained, striking a balance between adaptation and knowledge retention. In the following, we will present the prompt input design and the detailed architecture of \textsc{FLAG-Trader}.

\subsection{Prompt Input Design}

\begin{figure}[h]
    \centering
    \includegraphics[width=0.99\linewidth]{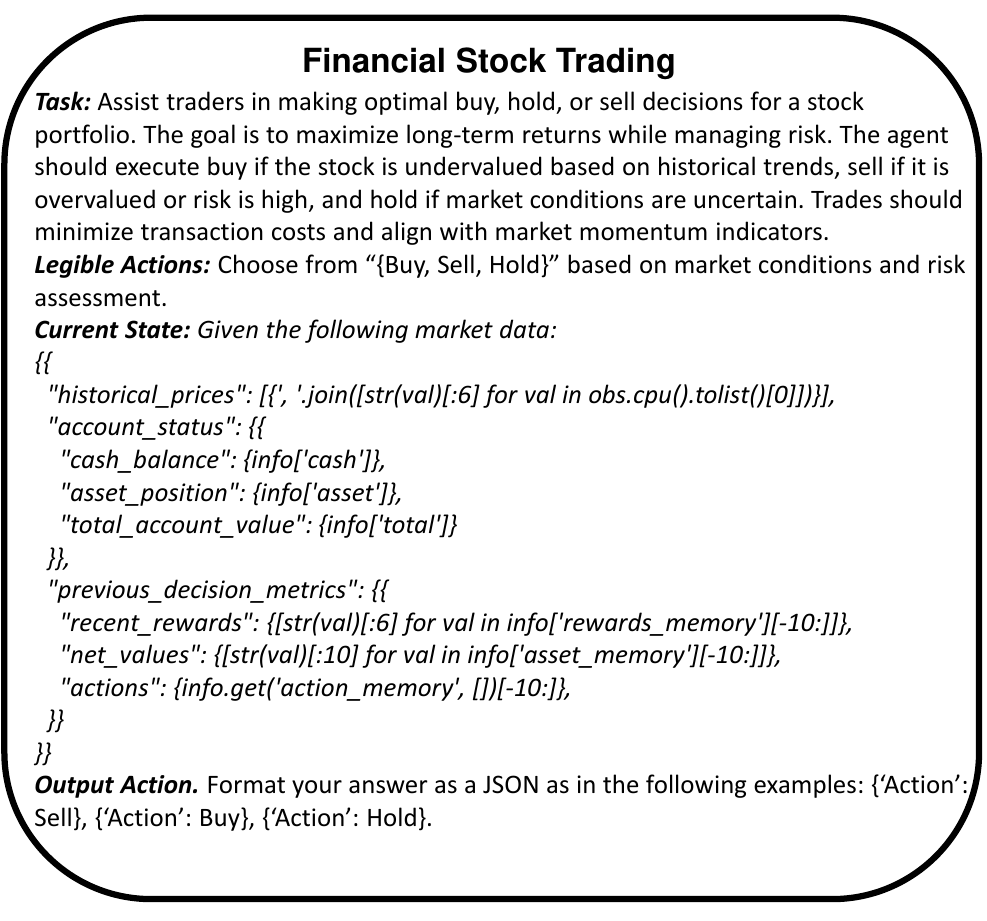}
    \caption{The format of input prompt. It contains the task description, the legible action set, the current state description, and the output action format.}
    \label{fig:prompt}
    \vspace{-0.2cm}
\end{figure}

The first stage of the pipeline involves designing a robust and informative prompt, denoted as \texttt{lang}($s_t$), which is constructed based on the current state 
$s_t$ to guide the LLM in making effective trading decisions. The prompt is carefully structured to encapsulate essential elements that provide context and ensure coherent, actionable outputs. It consists of four key components: a \emph{task description}, which defines the financial trading objective, outlining the problem domain and expected actions; a \emph{legible action space}, specifying the available trading decisions (\texttt{Sell,'' Hold,'' ``Buy''}); a \emph{current state representation}, incorporating market indicators, historical price data, and portfolio status to contextualize the decision-making process; and an \emph{output action}, which generates an executable trading decision. This structured prompt ensures that the LLM receives comprehensive input, enabling it to produce well-informed and actionable trading strategies, as illustrated in Figure \ref{fig:prompt}.

\subsection{\textsc{FLAG-Trader} Architecture}

To incorporate parameter-efficient fine-tuning into the policy gradient framework, we partition the intrinsic parameters of the LLM into two distinct components: the frozen parameters inherited from pretraining, denoted as 
$\theta_{\texttt{forzen}}$, and the trainable parameters, denoted as 
$\theta_{\texttt{train}}$. This separation allows the model to retain general language understanding while adapting to financial decision-making with minimal computational overhead.
Building upon this LLM structure, we introduce a policy network and a value network, both of which leverage the trainable top layers of the LLM for domain adaptation while sharing the frozen layers for knowledge retention. The overall architecture is illustrated in Figure \ref{fig:AC-network}.

\subsubsection{Policy Network Design}
The policy network is responsible for generating an optimal action distribution over the trading decision space $\cA$, conditioned on the observed market state. It consists of three main components:

\emph{State Encoding.}
To effectively process financial data using the LLM, the numerical market state 
$s$ is first converted into structured text using a predefined template\footnote{To simplify notation, we use \texttt{lang}($s_t$) to represent both the state encoding and the prompt, acknowledging this slight abuse of notation for convenience.}
\begin{align}
\texttt{lang}(s) = \text{"Price: \$}p\text{, Vol: }v\text{, RSI: }r\text{,..."}.
\label{eq:template}
\end{align}
This transformation enables the model to leverage the LLM’s textual reasoning capabilities, allowing it to understand and infer trading decisions in a structured, language-based manner.

\emph{LLM Processing.} The tokenized text representation of the state is then passed through the LLM backbone, which consists of:
1) \textbf{Frozen layers} (preserve general knowledge):
Token embeddings $E = \text{Embed}(\texttt{lang}(s))$ pass through LLM frozen layers, i.e.,
\begin{equation}
h^{(1)} = \text{LLM}_{1:N}(E;\theta_{\texttt{frozen}}).
\end{equation}
These layers preserve general knowledge acquired from pretraining, ensuring that the model maintains a strong foundational understanding of language and reasoning.
2) \textbf{Trainable layers} (domain-specific adaptation): The output from the frozen layers is then passed through the trainable layers, which are fine-tuned specifically for financial decision-making, i.e.,
\begin{equation}
h^{(2)} = \text{LLM}_{N+1:N+M}(h^{(1)};\theta_{\text{train}}).
\label{eq:trainable_layer}
\end{equation}
This structure enables efficient adaptation to the financial domain without modifying the entire LLM, significantly reducing training cost while maintaining performance.

\emph{Policy Head.} 
Finally, the processed representation is fed into the policy head, which outputs a probability distribution over the available trading actions according to
\begin{equation}
\texttt{logits} = \textsc{Policy\_Net}(h^{(2)},\theta_P)\in\mathbb{R}^{|\mathcal{A}|},
\label{eq:policy_dist}
\end{equation}
where $\theta_P$ is the parameter of \textsc{Policy\_Net},
with action masking for invalid trades: 
\begin{equation} 
\pi(a|s)\! =\!\! \begin{cases}
0 & a \notin \mathcal{A},\\
\frac{\exp(\texttt{logits}(a))}{\sum_{a'\in \mathcal{A}}\exp(\texttt{logits}(a^\prime))} & \text{otherwise}.
\end{cases}
\label{eq:masking}
\end{equation}
This ensures that actions outside the valid set 
$\cA$ (e.g., selling when no stocks are held) have zero probability, preventing invalid execution.

\subsubsection{Value Network Design}
The value network serves as the critic in the RL framework, estimating the expected return of a given state to guide the policy network's optimization. To efficiently leverage the shared LLM representation, the value network shares the same backbone as the policy network, processing the textual state representation through the frozen and trainable layers  \eqref{eq:template}–\eqref{eq:trainable_layer}. This design ensures efficient parameter utilization while maintaining a structured and informative state encoding. 
After passing through the LLM processing layers, the output 
$h^{(2)}$  is fed into a separate value prediction head, which maps the extracted features to a scalar value estimation:
\begin{equation}
V(s) = \textsc{Value\_Net}(h^{(2)}, \theta_V)\in\mathbb{R}^{1},
\label{eq:value_pred}
\end{equation}
where $\theta_V$ is the parameter of \textsc{Value\_Net}.

\subsection{Online Policy Gradient Learning}
The policy and value networks in \textsc{FLAG-Trader} are trained using an online policy gradient approach, ensuring that the model continuously refines its decision-making ability. The learning process follows an iterative cycle of state observation, action generation, reward evaluation, and policy optimization. The parameters of the model are updated using stochastic gradient descent (SGD), leveraging the computed policy and value losses to drive optimization.

At each training step, we define two key loss functions, i.e.,
\emph{policy loss} 
$\mathcal{L}_P$: measures how well the policy network aligns with the expected advantage-weighted log probability of actions; 
\emph{value loss} $\mathcal{L}_V$: ensures that the value network accurately estimates the expected return.

\begin{remark}
The definitions of \emph{policy loss} and \emph{value loss} may vary across different actor-critic (AC) algorithms. Here, we present a general formulation for clarity and ease of expression. Notably, our framework is designed to be flexible and adaptable, making it compatible with a wide range of AC algorithms.
\end{remark}

Based on these loss functions, the model updates the respective network parameters using backpropagation as follows. 

\textbf{Update Policy Head.} 
The policy network parameters 
 $\theta_P$ are updated via SGD to minimize the \emph{policy loss} $\mathcal{L}_P$ 
\begin{align}\label{eq:P}
    \theta_P\leftarrow\theta_P-\eta \nabla_{\theta_P}\mathcal{L}_P,
\end{align}
where $\eta$ is the learning rate for updating policy head $\theta_P$.

\textbf{Update Value Head.} The value network parameters 
$\theta_V$ are optimized via SGD to minimize the temporal difference (TD) error over policy loss $\mathcal{L}_V$
\begin{align}\label{eq:V}
    \theta_V\leftarrow\theta_V-\eta \nabla_{\theta_V}\mathcal{L}_V.
\end{align}

\textbf{Update Trainable LLM Layers.}
The trainable LLM parameters 
$\theta_{\texttt{train}}$ are updated via SGD jointly based on both the policy and value losses, i.e., $\mathcal{L}_P$ and $\mathcal{L}_V$, allowing the shared LLM representation to align with optimal decision-making:  
\begin{align}\label{eq:train}
    \theta_{\texttt{train}}\leftarrow\theta_{\texttt{train}}-\beta \nabla_{\theta_{\texttt{train}}}(\mathcal{L}_P+\mathcal{L}_V),
\end{align}
where $\beta$ is the learning rate for LLM parameter $\theta_{\texttt{train}}$. 

The updates in \eqref{eq:P}–\eqref{eq:train} are performed iteratively until the stopping criteria are met, as outlined in Algorithm \ref{alg:1}. This iterative learning process effectively balances exploration and exploitation, enhancing policy performance while maintaining stability. To mitigate overfitting and policy divergence, we employ Proximal Policy Optimization (PPO), which constrains updates by limiting the divergence from previous policies, ensuring more controlled and reliable learning. The detailed procedure of how to compute \emph{policy loss} $\cL_P$ and \emph{value loss} $\cL_P$ can be found in Appendix \ref{sec:Appendix_A}.

\begin{algorithm}[ht]
\caption{\textsc{FLAG-Trader}}
\begin{algorithmic}[1]
\STATE \textbf{Require:} Pre-trained LLM with parameter  $\theta:=(\theta_{\texttt{frozen}}, \theta_{\texttt{train}})$, environment dynamics $\cT$, reward function $\mathcal{R};$
\STATE Initialize policy network $\theta_P$ and value network $\theta_V$ with shared LLM trainable layers $\theta_{\texttt{train}}$;
\STATE Initialize experience replay buffer $B \leftarrow \emptyset$

\FOR{iteration $t=1,2,\ldots$,}
    \STATE Fetch the current state $s_t$ from the environment and construct an input prompt \texttt{lang}($s_t$);
    \STATE Pass prompt \texttt{lang}($s_t$) through LLM;
    \STATE \textsc{Policy\_Net} outputs $a_t$ from action space $\{\texttt{``buy,'' ``sell,'' ``hold''}\}$ based on \eqref{eq:masking};
    \STATE Execute action $a_t$ in the environment and observe reward $r(s_t, a_t)$ and transition to new state $s_{t+1}$;
    \STATE Store experience tuple $(s_t, a_t, r_t, s_{t+1})$ in replay buffer $B$;
    
    \IF{ $t \mod \tau=0$}
        \STATE Update policy head $\theta_P$ according to \eqref{eq:P};
        \STATE Update value head $\theta_V$ according to \eqref{eq:V};
        \STATE Update the trainable LLM layers $\theta_{\texttt{train}}$ according to \eqref{eq:train}.
        
    \ENDIF
\ENDFOR

\STATE \textbf{Return:} Fine-tuned \textsc{Policy\_Net}($\theta_P$).
\end{algorithmic}
\label{alg:1}
\end{algorithm}

\section{Experiments}
\begin{table*}[thbp]
\renewcommand{\arraystretch}{1}
\vspace{-0.2cm}
\setlength{\abovecaptionskip}{0.1cm}
\centering
\caption{Performance of stock trading with different LLMs as backbone model across seven stocks.}
\label{tab:stock-trading-performance1}
\begin{threeparttable}
\scalebox{0.70}{
\begin{tabular}{@{}lcccc|cccc|cccc@{}}
\toprule
\textbf{Model} & \multicolumn{4}{c}{\textbf{MSFT}} & \multicolumn{4}{c}{\textbf{JNJ}} & \multicolumn{4}{c}{\textbf{UVV}}\\
\cmidrule(lr){2-5}\cmidrule(lr){6-9}\cmidrule(lr){10-13}
& \textbf{CR\(\uparrow\)} & \textbf{SR\(\uparrow\)} & \textbf{AV\(\downarrow\)} & \textbf{MDD\(\downarrow\)}
& \textbf{CR\(\uparrow\)} & \textbf{SR\(\uparrow\)} & \textbf{AV\(\downarrow\)} & \textbf{MDD\(\downarrow\)}
& \textbf{CR\(\uparrow\)} & \textbf{SR\(\uparrow\)} & \textbf{AV\(\downarrow\)} & \textbf{MDD\(\downarrow\)}\\
\midrule
\textbf{Buy \& Hold}
& 15.340 & 1.039 & 24.980 & 9.428 
& 13.895 & 1.343 & 17.500 & 9.847 
& 36.583 & 2.112 & 29.299 & 15.406\\
\midrule
\multicolumn{13}{c}{\textit{\textbf{Financial Domain Models}}}\\
\textbf{Palmyra-Fin-70B}  
& 14.697 & 0.897 & 27.518 & 9.428 
& 5.748 & 0.450 & 19.317 & 9.367 
& 37.875 & 2.039 & 31.200 & 15.967\\
\midrule
\multicolumn{13}{c}{\textit{\textbf{Proprietary Models}}}\\
\textbf{GPT-o1-preview} 
& 17.184 & 0.962 & 30.000 & 9.428 
& 13.561 & 1.086 & 20.864 & 9.847 
& 41.508 & 2.147 & 32.479 & 9.633\\
\textbf{GPT-4} 
& 16.654 & 0.932 & 30.022 & 9.428 
& 13.712 & 1.103 & 20.894 & 9.860 
& 31.791 & 1.640 & 32.567 & 10.434\\
\textbf{GPT-4o} 
& 12.461 & 0.924 & 22.653 & 6.647
& 9.099 & 0.875 & 17.471 & 7.169 
& 8.043 & 0.496 & 27.241 & 14.889\\
\midrule
\multicolumn{13}{c}{\textit{\textbf{Open-Source Models}}}\\
\textbf{Qwen2.5-72B-Instruct}  
& 7.421 & 0.588 & 21.238 & 6.973 
& 14.353 & 1.140 & 20.995 & 9.812 
& 37.178 & 1.822 & 34.223 & 13.365\\
\textbf{Llama-3.1-70B-Instruct}  
& 17.396 & 1.335 & 21.892 & 7.045 
& 13.868 & 1.121 & 20.779 & 9.825 
& 35.981 & 1.728 & 34.986 & 15.406\\
\textbf{DeepSeek-67B-Chat} 
& 13.941 & 0.834 & 28.081 & 7.850 
& 14.426 & 1.185 & 20.450 & 9.825 
& 29.940 & 1.481 & 33.964 & 15.407\\
\textbf{Yi-1.5-34B-Chat}  
& 22.093 & 1.253 & 29.613 & 9.428 
& 14.004 & 1.180 & 19.938 & 9.847 
& 20.889 & 1.020 & 34.417 & 14.936\\
\textbf{Qwen2.5-32B-Instruct}  
& -0.557 & -0.041 & 22.893 & 8.946 
& 2.905 & 0.292 & 16.725 & 7.169 
& -1.623 & -0.097 & 27.973 & 17.986\\
\textbf{DeepSeek-V2-Lite (15.7B)}  
& 11.904 & 0.694 & 28.796 & 16.094 
& -7.482 & -0.670 & 18.773 & 17.806 
& 33.560 & 1.703 & 33.099 & 12.984\\
\textbf{Yi-1.5-9B-Chat}  
& 19.333 & 1.094 & 29.690 & 9.428 
& 18.606 & 1.611 & 19.409 & 10.986 
& 49.415 & 2.410 & 34.446 & 11.430\\
\textbf{Llama-3.1-8B-Instruct}  
& 22.703 & 1.322 & 28.855 & 7.385 
& 13.988 & 1.486 & 20.460 & 9.969 
& 41.108 & 1.981 & 34.866 & 16.429\\
\textbf{Qwen-2.5-Instruct-7B}  
& -10.305 & -0.724 & 23.937 & 23.371 
& 21.852 & 0.980 & 37.425 & 9.573 
& 11.752 & 0.853 & 22.988 & 15.451\\
\midrule
\multicolumn{13}{c}{\textit{\textbf{FLAG-TRADER }}}\\
\textbf{SmolLM2-135M-Instruct}  
& 20.106 & 1.373 & 24.932 & 9.428 
& 33.724 & 3.344 & 17.174 & 9.320
& 46.799 & 1.463 & 67.758 & 35.039\\
\bottomrule
\end{tabular}
}
\begin{tablenotes}
    \footnotesize
    \item[1] \small{The Buy \& Hold strategy is a passive investment approach commonly used as a baseline strategy, where an investor \\ purchases stocks and holds onto them for an extended period regardless of market fluctuations.}
    \item[2] \small{An upward arrow (\(\uparrow\)) next to a metric indicates that higher values signify better performance, while a downward arrow (\(\downarrow\))} \\
indicates that lower values are preferable.
    \item[3] \small{The numbers highlighted in red indicate the best-performing outcomes for the corresponding metrics.}
\end{tablenotes}
\end{threeparttable}
\vspace{-0.0cm}
\end{table*}

This section describes the overall experimental design and environmental setup for comparing the performance of different trading agents under consistent conditions.

\subsection{Experiment Setup}
For our single-asset trading tasks, we adopt two baselines: the buy-and-hold strategy and the LLM-based trading agent from \texttt{INVESTORBENCH} \cite{li2024investorbench}, which integrates 13 proprietary or open-source large language models. Our proposed model, \textsc{FLAG-Trader} (built on a 135M-parameter LLM), is then evaluated against these baselines for a comprehensive performance comparison.

We focus on five stocks and one crypto: Microsoft Corporation (MSFT), Johnson \& Johnson (JNJ), UVV Corporation (UVV), Honeywell International Inc. (HON), Tesla, Inc. (TSLA) and Bitcoin (BTC). As summarized in Table 1, each agent’s performance is measured across these assets. All language models use a temperature of 0.6 during inference to balance consistency and creativity in their responses.

We report four metrics-Composite Return (CR), Sharpe Ratio (SR), Annualized Volatility (AV), and Maximum Drawdown (MDD) and select final results from the test trajectory corresponding to the median of these metrics. If the median values arise from different epochs, we prioritize the run producing the median SR. Due to varying data availability, warm-up and test periods may differ. For the trading tasks of five stocks, the warm-up period is July 1, 2020, to September 30, 2020, and the test period is October 1, 2020, to May 6, 2021. On the other hand, the warm-up period of BTC trading is from 2023-02-11 to 2023-04-04 and the test period is from 2023-04-05 to 2023-11-05.

We deploy LLMs using the vllm framework, with configurations depending on model size. Small-scale models (under 10B parameters) run on two RTX A6000 GPUs (48GB each), mid-scale models (10B–65B parameters) require four RTX A6000 GPUs, and large-scale models (over 65B parameters) use eight A100 GPUs (80GB each). These setups provide sufficient resources for both inference and training, enabling a fair comparison of trading performance across different assets.
\textsc{FLAG-Trader} is trained by using PPO algorithm, which is detailed in Algorithm \ref{alg:flagtrader-ppo} in Appendix \ref{sec:Appendix_A}.

\subsection{Evaluation Metrics}
We use four widely recognized financial metrics \cite{hull2007risk} to evaluate and compare the investment performance of various LLM backbones across different tasks: Cumulative Return (CR), Sharpe Ratio (SR), Annualized Volatility (AV), and Maximum Drawdown (MDD). As CR and SR focus more on long-term gains and risk-adjusted returns, they are typically considered more important than AV and MDD for assessing asset trading performance. Accordingly, we treat CR and SR as our primary metrics for the final evaluation. 

\noindent\textbf{Cumulative Return (CR) \%} measures the total value change of an investment over time by summing logarithmic return calculated from daily PnL: 
\begin{align}
   \label{eq:cum_return}
   \textbf{CR} &= \sum_{t=1}^{T} \log{(1+\frac{pnl_t}{C_{t-1}+H_{t-1}P_{t-1}})},
\end{align}
where $pnl_t$ is the PnL at time $t$, and $C_{t-1}+H_{t-1}P_{t-1}$ is the account balance at time $t-1$. Notice that higher values indicate better strategy effectiveness.

\noindent\textbf{Sharpe Ratio (SR)} assesses risk-adjusted returns by dividing the average excess return ($R_p$) over the risk-free rate ($r_f$) by its volatility ($\sigma_p$):
\begin{equation}
    \textbf{SR} = \frac{R_p - r_f}{\sigma_p}.
\end{equation}  
Notice that higher ratios signify better performance.
  
\noindent\textbf{Annualized Volatility (AV) \% and Daily Volatility (DV) \%} quantify return fluctuations; AV is derived by scaling DV (\textit{standard deviation of daily logarithmic returns}) by the square root of the annual trading days (252): 
\begin{align}
   \label{eq:annuaVol}
    \textbf{AV} &= \textbf{DV} \times \sqrt{252}. 
\end{align} 
This metric highlights potential return deviations across the year.

\noindent\textbf{Max Drawdown (MDD) \%} calculates the largest  drop from peak to trough of the value of balance account:
    \begin{align}
    \label{eq:maxdrawdown}
    \textbf{MDD} = \text{max}(\frac{V_{\text{peak}} - V_{\text{trough}}}{V_{\text{peak}}}).
    \end{align}
Notice that  lower values indicate lesser risk and higher strategy robustness. 
\begin{table*}[thbp]
\renewcommand{\arraystretch}{1}
\vspace{-0.2cm}
\setlength{\abovecaptionskip}{0.1cm}
\centering
\caption{Performance of stock trading with different LLMs as backbone model across seven stocks.}
\label{tab:stock-trading-performance2}
\begin{threeparttable}
\scalebox{0.68}{\begin{tabular}{@{}lcccc|cccc|cccc@{}}
\toprule
\textbf{Model} & \multicolumn{4}{c}{\textbf{HON}} &\multicolumn{4}{c}{\textbf{TSLA}} & \multicolumn{4}{c}{\textbf{BTC}} \\
\cmidrule(lr){2-5}\cmidrule(lr){6-9}\cmidrule(lr){10-13}
& \textbf{CR\(\uparrow\)} & \textbf{SR\(\uparrow\)} & \textbf{AV\(\downarrow\)} & \textbf{MDD\(\downarrow\)}
& \textbf{CR\(\uparrow\)} & \textbf{SR\(\uparrow\)} & \textbf{AV\(\downarrow\)} & \textbf{MDD\(\downarrow\)}
& \textbf{CR\(\uparrow\)} & \textbf{SR\(\uparrow\)} & \textbf{AV\(\downarrow\)} & \textbf{MDD\(\downarrow\)} \\
\midrule
\textbf{Buy \& Hold}

& 33.256 & 2.347 & 23.967 & 9.195 
& 39.244 & 0.869 & 75.854 & 37.975
& 21.821 & 0.683 & 37.426 & 20.796 \\
\midrule
\multicolumn{13}{c}{\textit{\textbf{Financial Domain Models}}} \\
\textbf{Palmyra-Fin-70B}  

& 20.016 & 1.464 & 22.974 & 6.824 
& -6.661 & -0.222 & 50.379 & 25.820

& -20.812 & -1.212 & 20.036 & 27.782\\
\midrule
\multicolumn{13}{c}{\textit{\textbf{Proprietary Models}}} \\
\textbf{GPT-o1-preview} 

& 13.162 & 0.776 & 28.511 & 11.558 
& 34.499 & 0.796 & 72.822 & 35.490

& 34.060 & 1.114 & 35.846 & 17.075\\
\textbf{GPT-4} 

& 34.342 & 2.005 & 28.779 & 9.195
& 45.246 & 1.190 & 63.896 & 25.031

& 22.396 & 0.828 & 31.699 & 17.206\\
\textbf{GPT-4o} 
 
& 38.540 & 2.418 & 26.782 & 8.979 
& 45.946 & 1.348 & 57.281 & 21.631

& 14.330 & 0.532 & 31.304 & 17.278\\
\midrule
\multicolumn{13}{c}{\textit{\textbf{Open-Source Models}}} \\
\textbf{Qwen2.5-72B-Instruct}  

& 34.309 & 2.000 & 28.779 & 9.292 
& 39.112 & 1.075 & 61.136 & 26.985

& 0.549 & 0.325 & 1.979 & 0.897\\
\textbf{Llama-3.1-70B-Instruct}  

& 43.944 & 2.646 & 27.903 & 8.993 
& 37.545 & 0.891 & 70.815 & 29.813

& 20.440 & 0.758 & 31.604 & 17.813\\
\textbf{DeepSeek-67B-Chat} 

& 32.536 & 1.909 & 28.628 & 10.782 
& 35.647 & 0.885 & 67.660 & 33.359

& 28.307 & 0.891 & 37.219 & 17.944\\
\textbf{Yi-1.5-34B-Chat}  

& 30.743 & 1.823 & 28.335 & 9.195 
& 35.364 & 0.808 & 73.561 & 35.490

& 13.620 & 0.434 & 36.778 & 22.790 \\
\textbf{Qwen2.5-32B-Instruct}  

& 26.332 & 1.980 & 22.348 & 5.261
& 21.336 & 0.729 & 49.157 & 20.704

& 11.566 & 0.869 & 15.608 & 7.984\\
\textbf{DeepSeek-V2-Lite (15.7B)}  

& 16.686 & 0.974 & 28.771 & 16.806
& 31.458 & 0.744 & 68.524 & 35.404

& 4.804 & 0.153 & 36.846 & 20.562\\
\textbf{Yi-1.5-9B-Chat}  

& 29.028 & 1.700 & 28.682 & 12.588 
& 31.350 & 0.703 & 74.895 & 37.975

 & 7.953 & 0.253 & 36.799 & 26.545 \\
\textbf{Llama-3.1-8B-Instruct}  

& 39.079 & 2.320 & 28.299 & 10.341
& 35.622 & 0.832 & 71.936 & 36.383

& 20.521 & 0.646 & 37.240 & 21.104\\
\textbf{Qwen-2.5-Instruct-7B}  

& 4.291 & 0.285 & 24.933 &14.156 
& 41.203 & 0.925 & 74.862 & 37.975

& 19.477 & 0.612 & 37.289 & 20.796\\
\midrule
\multicolumn{13}{c}{\textit{\textbf{FLAG-TRADER }}} \\
\textbf{SmolLM2-135M-Instruct}  
 
& 34.342 & 2.429 & 23.913 & 10.872
& 50.394 & 1.362 & 64.004 & 37.975

& 45.511 & 1.734 & 30.903 & 24.440\\
\bottomrule
\end{tabular}
}
\begin{tablenotes}
    \footnotesize
    \item[1] \small{The Buy \& Hold strategy is a passive investment approach commonly used as a baseline strategy, where an investor \\ purchases stocks and holds onto them for an extended period regardless of market fluctuations.}
    \item[2] \small{An upward arrow (\(\uparrow\)) next to a metric indicates that higher values signify better performance, while a downward arrow (\(\downarrow\))} \\
indicates that lower values are preferable.
    \item[3] \small{The numbers highlighted in red indicate the best-performing outcomes for the corresponding metrics.}
\end{tablenotes}
\end{threeparttable}
\vspace{-0.0cm}
\end{table*}
\subsection{Experimental Results}
\begin{itemize}
    \item \noindent\textbf{FLAG-Trader achieves superior stock trading performance.} Unlike the baseline agent, which relies on an LLM-agentic framework, \textsc{FLAG-Trader} undergoes an RL-based post-training phase. As shown in Table~\ref{tab:stock-trading-performance1} and Table \ref{tab:stock-trading-performance2}, \textsc{FLAG-Trader} consistently outperforms the baseline across multiple metrics, demonstrating its stronger adaptability and optimization in diverse market environments.

    \item \textbf{Convergence to a (stable) optimal policy.} When using an LLM as the policy network in deep RL, the policy is jointly parameterized by the model’s intrinsic parameters and the prompt. Although the initial prompts strongly influence policy generation during the first few training epochs, this effect diminishes over time. Consequently, the system converges to a relatively stable policy that becomes less sensitive to the initial prompts, indicating that RL training allows the LLM-based agent to refine its strategy and achieve a robust trading policy.

    \item \noindent\textbf{\textsc{FLAG-Trader} enables small-scale models to surpass large-scale counterparts.} While increasing model size generally enhances financial decision-making and robustness— as seen with large proprietary models (e.g., GPT-o1-preview) in the baseline framework-\textsc{FLAG-Trader} leverages an RL-based training pipeline to enable a 135M-parameter open-source model to outperform significantly larger models in financial trading tasks. This demonstrates that a well-designed training strategy can bridge or even surpass the performance gap typically associated with model scale.

\end{itemize}

\section{Conclusion}
In this paper, we introduced \textsc{FLAG-Trader}, a novel framework that integrates LLMs with RL for financial trading. In particular, \textsc{FLAG-Trader} leverages LLMs as policy networks, allowing for natural language-driven decision-making while benefiting from reward-driven optimization through RL fine-tuning. Our framework enables small-scale LLMs to surpass larger proprietary models by efficiently adapting to market conditions via a structured reinforcement learning approach. Through extensive experiments across multiple stock trading scenarios, we demonstrated that \textsc{FLAG-Trader} consistently outperforms baseline methods, including LLM-agentic frameworks and conventional RL-based trading agents. These results highlight the potential of integrating LLMs with RL to achieve adaptability in financial decision-making.

\newpage
\section*{Limitations and Potential Risk}

Despite its promising results, \textsc{FLAG-Trader} has several limitations. First, while our approach significantly enhances the decision-making ability of LLMs, it remains computationally expensive, particularly when fine-tuning on large-scale market datasets. Reducing computational overhead while maintaining performance is an important direction for future research. Second, financial markets exhibit high volatility and non-stationarity, posing challenges for long-term generalization. Future work should explore techniques such as continual learning or meta-learning to enhance model adaptability in evolving market conditions. Third, while \textsc{FLAG-Trader} effectively integrates textual and numerical data, its reliance on structured prompts could introduce biases in decision-making. Improving prompt design or exploring retrieval-augmented methods may further enhance robustness. Lastly, real-world trading requires stringent risk management, and \textsc{FLAG-Trader} currently optimizes for financial returns without explicitly incorporating risk-sensitive constraints. Extending the framework to integrate risk-aware objectives and dynamic portfolio optimization could provide more robust and practical financial trading solutions.



\bibliography{main}

\newpage
\appendix
\onecolumn
\section{Additional Algorithmic  Details: \textsc{FLAG-Trader} with PPO}\label{sec:Appendix_A}

In this section, we outline a detailed procedure for training the \textsc{FLAG-Trader} architecture via PPO, where the \textsc{Policy\_Net} (actor) and the \textsc{Value\_Net} (critic) share a subset of trainable parameters from a LLM, with
$\theta = \big( \theta_{\texttt{train}}, \theta_P, \theta_V\big)$.
We define $\theta_{policy} = \big( \theta_{\texttt{train}}, \theta_P)$ and $\theta_{value} = \big( \theta_{\texttt{train}}, \theta_V)$ for simplicity.

\textbf{Advantage Estimation.}
We use the Generalized Advantage Estimation (GAE) to compute the advantage function \( A_t \):
\begin{align}
    A_t \;=\; \sum_{k=0}^{T-1} (\gamma \lambda)^k \bigl[r_{t+k} + \gamma V_{\theta_{value}}(s_{t+k+1}) - V_{\theta_{value}}(s_{t+k})\bigr],
\end{align}
where \( \gamma \) is the discount factor, and \( \lambda \) is the GAE parameter.

\textbf{Probability Ratio.}
Let \(\theta_{policy, \mathrm{old}}\) denote the parameters before the current update. The PPO probability ratio is
\begin{align}
    r_t(\theta_{policy}) \;=\; \frac{\pi_{\theta_{policy}}(a_t \mid s_t)}{\pi_{\theta_{policy,\mathrm{old}}}(a_t \mid s_t)}.
\end{align}

\textbf{PPO Clipped Objective.}
PPO clips this ratio to prevent overly large updates. The surrogate objective is
\begin{align}
    \mathcal{L}_P(\theta_{policy}) \;=\; \mathbb{E}_t \Bigl[
  \min\bigl(r_t(\theta_{policy})\,A_t,\; \text{clip}\bigl(r_t(\theta_{policy}),\,1-\varepsilon,\,1+\varepsilon\bigr)\,A_t\bigr)
\Bigr],
\end{align}
where \(\varepsilon\) is a hyperparameter.

\textbf{Value Function Loss.}
The critic (value network) is updated by minimizing the difference between the predicted value \(V_{\theta_{value}}(s_t)\) and the target return \(R_t\). A common choice is:
\begin{align}
    \mathcal{L}_{V}(\theta_{value}) \;=\; \mathbb{E}_t\Bigl[(V_{\theta_{value}}(s_t) - R_t)^2\Bigr].
\end{align}

\textbf{Combined Loss.}
We often add an entropy term to encourage exploration, yielding the overall objective:
\begin{align}
    \mathcal{L}_{\text{total}}(\theta)
\;=\;
-\,\mathcal{L}_{P}(\theta_{policy})
\;+\;
c_1\,\mathcal{L}_{V}(\theta_{value})
\;-\;
c_2\,\mathcal{H}\bigl(\pi_{\theta_{policy}}\bigr),
\end{align}
where \(c_1\) and \(c_2\) are weighting coefficients, and \(\mathcal{H}(\pi_{\theta_{policy}})\) represents the policy entropy.

\textbf{Parameter Updates.}
At each iteration, we apply gradient descent on the total loss:
\begin{align}
\theta_P       &\leftarrow \theta_P       \;-\; \eta \;\nabla_{\theta_P}\,\mathcal{L}_P, \\
\theta_V       &\leftarrow \theta_V       \;-\; \eta \;\nabla_{\theta_V}\,\mathcal{L}_V, \\
\theta_{\texttt{train}} &\leftarrow \theta_{\texttt{train}} - \beta \;\nabla_{\theta_{\text{train}}}\,\mathcal{L}_{\text{total}},
\end{align}
where \(\eta\) and \(\beta\) are learning rates for the policy head, value head, and trainable LLM layers respectively. The algorithm is summerized in Algorithm \ref{alg:flagtrader-ppo}.

\begin{algorithm}[H]
\caption{FLAG-TRADER with PPO}
\label{alg:flagtrader-ppo}
\begin{algorithmic}[1]
\STATE \textbf{Input:} Pre-trained LLM parameters $(\theta_{\texttt{frozen}}, \theta_{\texttt{train}})$; actor parameters $\theta_P$; critic parameters $\theta_V$; environment $\mathcal{E}$; discount factor $\gamma$; GAE parameter $\lambda$; PPO clip $\varepsilon$; learning rates $\eta, \beta$;
\STATE Initialize $\theta_{\text{train}}, \theta_P, \theta_V$; let $\theta_{\mathrm{old}} \leftarrow \theta$
\STATE Initialize replay buffer $B \leftarrow \emptyset$
\FOR{iteration = 1 to \text{max\_iters}}
  \STATE // \textit{Collect Rollouts}
  \FOR{t = 1 to T}
    \STATE Fetch the current state $s_t$ from the environment and construct an input prompt \texttt{lang}($s_t$);
    \STATE Pass prompt \texttt{lang}($s_t$) through LLM;
    \STATE \textsc{Policy\_Net} outputs $a_t$ from action space $\{\texttt{``buy,'' ``sell,'' ``hold''}\}$ based on \eqref{eq:masking};
    \STATE Execute action $a_t$ in the environment and observe reward $r(s_t, a_t)$ and transition to new state $s_{t+1}$;
    \STATE Store experience tuple $(s_t, a_t, r_t, s_{t+1})$ in replay buffer $B$;
    
  \ENDFOR

  \STATE // \textit{Compute Advantage and Targets}
  \FOR{\textbf{each} transition in $B$}
    \STATE Compute $V_{\theta_{value}}(s_t)$ and advantage $A_t$ (e.g., via GAE)
  \ENDFOR

  \STATE // \textit{Perform PPO Updates}
  \FOR{update\_epoch = 1 to K}
    \STATE Sample mini-batch $\mathcal{M}$ from $B$
    \STATE Compute probability ratio $r_t(\theta_{policy}) \;=\; \frac{\pi_{\theta_{policy}}(a_t \mid s_t)}{\pi_{\theta_{policy,\mathrm{old}}}(a_t \mid s_t)}$;
    \STATE Compute PPO loss $\mathcal{L}_P(\theta_{policy}) \;=\; \mathbb{E}_t \Bigl[
  \min\bigl(r_t(\theta_{policy})\,A_t,\; \text{clip}\bigl(r_t(\theta_{policy}),\,1-\varepsilon,\,1+\varepsilon\bigr)\,A_t\bigr)
\Bigr]$;
    \STATE Compute Value loss $\mathcal{L}_{V}(\theta_{value}) \;=\; \mathbb{E}_t\Bigl[(V_{\theta_{value}}(s_t) - R_t)^2\Bigr]$;
    \STATE Compute total loss $\mathcal{L}_{\text{total}}(\theta)
\;=\;
-\,\mathcal{L}_{P}(\theta_{policy})
\;+\;
c_1\,\mathcal{L}_{V}(\theta_{value})
\;-\;
c_2\,\mathcal{H}\bigl(\pi_{\theta_{policy}}\bigr)$;
    \STATE Perform gradient descent on each parameter group:
    \begin{align*}
\theta_P       &\leftarrow \theta_P       \;-\; \eta \;\nabla_{\theta_P}\,\mathcal{L}_P, \\
\theta_V       &\leftarrow \theta_V       \;-\; \eta \;\nabla_{\theta_V}\,\mathcal{L}_V, \\
\theta_{\texttt{train}} &\leftarrow \theta_{\texttt{train}} - \beta \;\nabla_{\theta_{\text{train}}}\,\mathcal{L}_{\text{total}};
\end{align*}
  \ENDFOR

  \STATE // \textit{Update old policy parameters}
  \STATE Update $\theta = \big( \theta_{\texttt{train}}, \theta_P, \theta_V\big)$ by $\theta_{\mathrm{old}} \leftarrow \theta$;
\ENDFOR
\STATE \textbf{Return:} Fine-tuned \textsc{Policy\_Net}($\theta_P$).
\end{algorithmic}
\end{algorithm}

\section{Additional Experimental Details}

\subsection*{Hyperparameters for Finetuening \textsc{FLAG-Trader} with PPO in Algorithm \ref{alg:flagtrader-ppo}}
\begin{table}[!ht]
\centering
\caption{\textsc{FLAG-Trader} with PPO Finetuning Hyperparameters and Settings.}
\label{tab:parameters:lora}
\scalebox{0.8}{
\begin{tabular}{lll}
\toprule
\textbf{Parameter} & \textbf{Default Value} & \textbf{Description} \\
\toprule
\texttt{total\_timesteps} & 13860 & Total number of timesteps \\
\texttt{learning\_rate} & \(5 \times 10^{-4}\) & Learning rate of optimizer \\
\texttt{num\_envs} & 1 & Number of parallel environments \\
\texttt{num\_steps} & 40 & Steps per policy rollout \\
\texttt{anneal\_lr} & True & Enable learning rate annealing \\
\texttt{gamma} & 0.95 & Discount factor \(\gamma\) \\
\texttt{gae\_lambda} & 0.98 & Lambda for Generalized Advantage Estimation \\
\texttt{update\_epochs} & 1 & Number of update epochs per cycle \\
\texttt{norm\_adv} & True & Advantages whitening \\
\texttt{clip\_coef} & 0.2 & Surrogate clipping coefficient \\
\texttt{clip\_vloss} & True & Clipped loss for value function \\
\texttt{ent\_coef} & 0.05 & Coefficient of entropy term \\
\texttt{vf\_coef} & 0.5 & Coefficient of value function \\
\texttt{kl\_coef} & 0.05 & KL divergence with reference model \\
\texttt{max\_grad\_norm} & 0.5 & Maximum gradient clipping norm \\
\texttt{target\_kl} & None & Target KL divergence threshold \\
\texttt{dropout} & 0.0 & Dropout rate \\
\texttt{llm} & "SmolLM2-135M-Instruct" & Model to fine-tune \\
\texttt{train\_dtype} & "float16" & Training data type \\
\texttt{gradient\_accumulation\_steps} & 8 & Number of gradient accumulation steps \\
\texttt{minibatch\_size} & 32 & Mini-batch size for fine-tuning \\
\texttt{max\_episode\_steps} & 65 & Maximum number of steps per episode \\
\bottomrule
\end{tabular}
}
\end{table}

\end{document}